\documentclass[letterpaper]{article}

\usepackage{arxiv}

\usepackage[utf8]{inputenc} 
\usepackage[T1]{fontenc}    
\usepackage[unicode]{hyperref}       
\usepackage{url}            
\usepackage{booktabs}       
\usepackage{amsfonts}       
\usepackage{nicefrac}       
\usepackage{microtype}      
\usepackage{lipsum}		
\usepackage{graphicx}
\usepackage{natbib}
\usepackage{doi}
\usepackage{subcaption}
\usepackage[scaled=0.86]{helvet}

\PassOptionsToPackage{hyphens}{url}

\newcommand{\thetitle}{The \texttt{Explabox}: Model-Agnostic Machine Learning Transparency \& Analysis}
\title{\thetitle}
\renewcommand{\shorttitle}{\thetitle}
\date{}  

\author{ \href{https://orcid.org/0000-0002-6430-9774}{\includegraphics[scale=0.06]{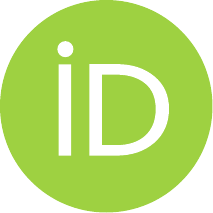}\hspace{1mm}Marcel Robeer}$^{1,2}$\thanks{Corresponding author: \href{mailto:m.j.robeer@uu.nl}{m.j.robeer@uu.nl}.} \quad \href{https://orcid.org/0000-0002-4823-6085}{\includegraphics[scale=0.06]{orcid.pdf}\hspace{1mm}Michiel Bron}$^{1,2}$ \quad \href{https://orcid.org/0000-0002-2729-6599}{\includegraphics[scale=0.06]{orcid.pdf}\hspace{1mm}Elize Herrewijnen}$^{1,2}$ \quad Riwish Hoeseni$^2$ \quad \href{https://orcid.org/0000-0002-5699-9656}{\includegraphics[scale=0.06]{orcid.pdf}\hspace{1mm}Floris Bex}$^{1,3}$\\
\\
$^1$National Police Lab AI, Utrecht University, The Netherlands\\
$^2$Netherlands National Police, The Netherlands\\
$^3$Tilburg Institute for Law, Technology and Society, Tilburg University, The Netherlands
}

\hypersetup{
    pdftitle={\thetitle},
    pdfsubject={cs.AI, stat.ML, cs.LG},
    pdfauthor={Marcel Robeer, Michiel Bron, Elize Herrewijnen, Riwish Hoeseni, and Floris Bex},
    pdfkeywords={Explainable AI (XAI), Interpretability, Fairness, Robustness, Safety, Auditability},
    colorlinks=true,
    allcolors=blue,
}

\begin{document}
\maketitle
\vspace*{-1em}
\begin{abstract}
We present the \texttt{Explabox}: an open-source toolkit for transparent and responsible machine learning (ML) model development and usage. \texttt{Explabox} aids in achieving explainable, fair and robust models by employing a four-step strategy: \textit{explore}, \textit{examine}, \textit{explain} and \textit{expose}. These steps offer model-agnostic analyses that transform complex `ingestibles' (models and data) into interpretable `digestibles'. The toolkit encompasses digestibles for descriptive statistics, performance metrics, model behavior explanations (local and global), and robustness, security, and fairness assessments. Implemented in Python, \texttt{Explabox} supports multiple interaction modes and builds on open-source packages. It empowers model developers and testers to operationalize explainability, fairness, auditability, and security. The initial release focuses on text data and models, with plans for expansion. \texttt{Explabox}'s code and documentation are available open-source at \href{https://explabox.readthedocs.io/en/stable}{https://explabox.readthedocs.io}.
\end{abstract}


\section{Introduction}
It is crucial that Machine Learning (ML) development and usage is done in a responsible and transparent manner. High-stakes organizational decisions may significantly impact individuals and society, with potential severe consequences stemming from biases or model errors. This is exemplified by the EU AI Act's regulatory framework, requiring high-risk systems to be properly tested, documented and assessed on their conformity before being applied in practice \citep{Edwards2022}. Yet, operationalizing transparency (explainable ML) and testing model behavior (fairness, robustness, and security auditing) remains a difficult and laborious task, given the myriad of techniques and their associated learning curves. In response, we have devised a comprehensive four-step analysis strategy---\textit{explore}, \textit{examine}, \textit{explain}, and \textit{expose}---ensuring holistic model transparency and testing. The open-source \texttt{Explabox} offers these analyses through well-documented, reproducible steps. Data scientists can now access a unified, model-agnostic approach developed and used in a high-stakes context---the Netherlands National Police---that is applicable to any text classifier or regressor.

We have developed the \texttt{Explabox} in an organizational environment where models and data are analyzed repeatedly, and where internal and external stakeholders have varying explanatory needs and preferred formats. Several related tools have been made available, such as \texttt{AIX360} \citep{Arya2019}, \texttt{alibi} explain \citep{Klaise2021}, \texttt{dalex} \citep{Baniecki2021}, \texttt{CheckList} \citep{Ribeiro2020}, and \texttt{AIF360} \citep{Bellamy2018}. However, these tools exhibit shortcomings such as incompatibility with recent Python versions (3.8--3.12), restricted software functionality primarily focused on testing or explainability, an absence of reproducible outcomes, or that they do not provide the flexibility regarding how results can be communicated to address stakeholder needs. To fill this gap, we propose the \texttt{Explabox}. The \texttt{Explabox} is an open-source Python toolkit that supports organizations with responsible ML development with minimal disruption to practitioners' workflows through a four-step analysis strategy easily embedded with existing datasets and models, and providing a central node for connecting with state-of-the-art research and sharing best practices.

\section{The \texttt{Explabox}: Explore, Examine, Explain \& Expose your ML models}
The \texttt{Explabox} transforms opaque \textit{ingestibles} into transparent \textit{digestibles} through four types of \textit{analyses}. The digestibles provide insights into model behavior and data, enhancing model explainability and assisting in auditing the fairness, robustness, and security of ML systems.

\begin{figure}[!htb]
    \begin{subfigure}[b]{0.5\textwidth}
        \centering
        \includegraphics{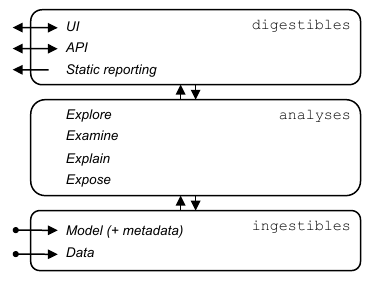}
        \caption{Logical separation into layers with interfaces}
        \label{fig:layers}
    \end{subfigure}
    \begin{subfigure}[b]{0.5\textwidth}
        \centering
        \includegraphics{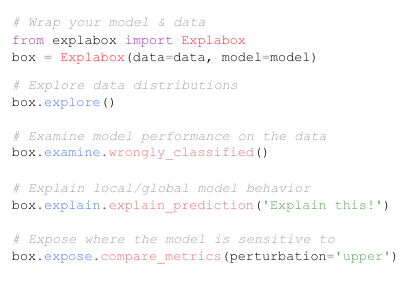}
        \caption{Python snippet with example analyses}
        \label{fig:Python}
    \end{subfigure}
    \par\smallskip 
    \begin{subfigure}[b]{\textwidth}
            \centering
        \includegraphics[width=\textwidth]{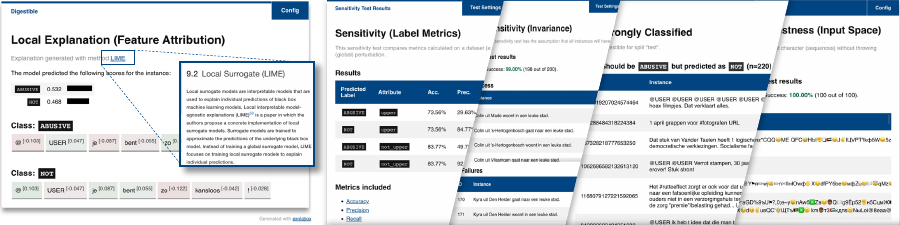}
        \caption{UI elements of the Jupyter Notebook interface for interactive explainability and analyses}
        \label{fig:ui}
    \end{subfigure}
    \caption{The \texttt{Explabox} facilitates model-agnostic responsible AI development with explainability and versatile analysis tools, accessible through user-friendly interfaces.}
    \label{fig:explabox}
\end{figure}

\subsection{Ingestibles}
Ingestibles serve as a unified interface for importing models and data. The layers (Fig~\ref{fig:layers}) abstract away from how the model and data are accessed, and allow for optimized processing. The \texttt{Explabox} encapsulates the model and data with \texttt{instancelib} \citep{instancelib} to ensure fast processing. The model can be any Python \texttt{Callable} containing a regression or (binary and multi-class) classification model. Models developed with \texttt{scikit-learn} \citep{scikit-learn} or with inferencing through \texttt{onnx} (e.g., PyTorch and TensorFlow/Keras) can be imported directly with further optimizations and automatic extraction of how inputs/outputs are to be interpreted. Data can be automatically downloaded, extracted and loaded. Data can be provided as \texttt{NumPy} arrays, \texttt{Pandas} DataFrames, \texttt{huggingface} datasets, raw files (e.g., HDF5, CSV or TSV), or as (compressed) folders containing raw files. The data can be subdivided into named splits (e.g., train-test-validation), and instance vectors and tokens can be precomputed (and optionally saved on disk) to provide fast inferencing.

\subsection{Analyses}
The \texttt{Explabox} turns these \textit{ingestibles} into \textit{digestibles}: pieces of information that increase the transparency of the ingestibles. Turning ingestibles into digestibles is done through four types of analyses: \textsf{explore}, \textsf{examine}, \textsf{explain} and \textsf{expose}.

\paragraph{\textsf{Explore}}\hspace*{-1em} 
allows slicing, dicing and sorting data, and provides descriptive statistics, grouped by named split. Relevant statistics include data set sizes and label distributions, and modality-relevant information, such as string lengths and tokenized lengths for textual data.

\paragraph{\textsf{Examine}}\hspace*{-1em}
shows performance metrics of the ML model on the data, summarized in a table or shown graphically. Metrics are accompanied by references on how they are computed and how they should be interpreted. For further analysis, \textsf{examine} also supports to drill-down into which instances were predicted correctly and incorrectly.

\paragraph{\textsf{Explain}}\hspace*{-1em}
uses model-agnostic techniques \citep{Ribeiro2016b} to explain model behavior (\textit{global}) and individual predictions (\textit{local}). It summarizes model-labelled data, through prototypes (\texttt{K-Medoids}) or prototypes and criticisms \citep[\texttt{MMDCritic}; ][]{Kim2016}, and token distributions (\texttt{TokenFrequency}) and informativeness (\texttt{TokenInformation}) for the text modality. Local explanations are given by popular techniques for feature attribution scores [\texttt{LIME} \citep{Ribeiro2016a}, \texttt{KernelSHAP} \citep{Lundberg2017}], relevant feature subsets [\texttt{Anchors} \citep{Ribeiro2018}], local rule-based models [\texttt{LORE} \citep{Guidotti2018}], and counterfactual/contrastive explanations [\texttt{FoilTrees} \citep{Waa2018}]. These methods are constructed from generic components that split the relevant steps in global and local explanation generation. This allows customizability and for scientific advancements to quickly end up in operational processes. For example, they may combine the data sampling from \texttt{KernelSHAP} and summarize these with a surrogate rule-based model provided by the \texttt{imodels} package \citep{Singh2021}. However, to ease users into adoption we provide example configurations, such as \texttt{LIME} with default hyperparameters. \textsf{Explain} is provided by subpackage \texttt{text\_explainability} \citep{text_explainability}, which doubles as a standalone tool.

\paragraph{\textsf{Expose}}\hspace*{-1em}
gathers sensitivity insights through local/global testing regimes. These insights can be used to, through relevant attributes, assess the \textit{robustness} (e.g., the effect of typos on model performance), \textit{security} (e.g., if inputs containing certain characters crash the model) and \textit{fairness} (e.g., subgroup performance for protected attributes such as country of origin, gender, race or socioeconomic status) of the model. Relevant attributes can either be observed in the current data or generated from user-provided templates \citep{Ribeiro2020}. These attributes are then either summarized in performance metrics, compared to expected behavior \citep{Ribeiro2020}, or assessed with fairness metrics for classification \citep{Mehrabi2021} and regression \citep{Agarwal2019}. Like \textsf{explain}, \textsf{expose} is also made from generic components, which allows users to customize data generation and tests. \textsf{Expose} is provided by the \texttt{text\_sensitivity} subpackage \citep{text_sensitivity}, which also doubles as a standalone tool.

\subsection{Digestibles}
To serve diverse stakeholders' needs---such as auditors, applicants, end-users or clients \citep{Tomsett2018}---in consuming model and data insights, the digestibles are accessible through different channels like an interactive user interface (UI) for Jupyter Notebook or webpages (Fig~\ref{fig:ui}), an API for integration with other tooling, and static reporting.

\subsection{Open-source implementation}
\texttt{Explabox} is a Python library with full cross-platform support for versions 3.8--3.12 (see Fig.~\ref{fig:Python} for example analyses). Distributed under the GNU LGPL-3.0 license, it offers a flexible inferencing and data-handling API through \texttt{instancelib} \citep{instancelib}.  
It also benefits from other open-source communities, such as \texttt{scikit-learn} \citep{scikit-learn} for surrogate models, clustering and dimensionality reduction, \texttt{imodels} \citep{Singh2021} for rule-based interpretable models, \texttt{Faker} \citep{Faker} for multi-language data generation, and \texttt{plotly} \citep{plotly} for interactive and static graphics.

The \texttt{Explabox} documentation provides installation guides, a tutorial to get started, a comprehensive example use-case using Jupyter Notebook, a full overview of all Python classes and functions, and guides on how to contribute. Documentation generation, testing, code quality assurance, and versioning are automated with a GitHub CI/CD pipeline.

\section{Conclusion and Future Work}
The \texttt{Explabox} enables organizations to responsibly develop and apply AI applications. The model-agnostic toolkit provides interactive analyses that model developers and testers can use to operationalize, report and discuss model explainability, fairness, auditability, and security. It provides flexibility to handle various ML use-cases for classification and regression, while also being a central node for standardization and connecting research with practice. The first full release, focusing on text data and models, is available open-source and fully documented at \href{https://explabox.readthedocs.io/en/stable/}{https://explabox.readthedocs.io}.

Further development has already started, where we focus on extending the \texttt{Explabox} for (1) models and data with the tabular, audio and image modality, and multi-modal combinations; (2) improved integration of data and model provenance for auditability; (3) comparisons of multiple datasets and multiple models, and; (4) research and development for a new interface and improved stakeholder data visualization.

\section*{Acknowledgements}
Development of the Explabox was supported by the Netherlands National Police. The authors would like to thank all anonymous contributors within the Netherlands National Police for their contributions to software development, testing and active usage. In addition, the authors would like to thank the participants of the demos at ICT.OPEN 2022, the National Police Lab AI at Utrecht University, and the University Medical Center Utrecht for their valuable feedback.

\bibliographystyle{plainnat}
\bibliography{references}

\end{document}